\documentclass[preprint,12pt]{elsarticle}

\journal{arXiv}

\usepackage{comment}
\usepackage[autonum,colorhypersetup]{shortex} 
\usepackage{thmtools, thm-restate}
\crefname{appendix}{}{}

\usepackage{pgfplots}
\pgfplotsset{compat=newest} 
\usetikzlibrary{arrows.meta}
\usetikzlibrary{backgrounds}
\usepgfplotslibrary{patchplots}
\usepgfplotslibrary{groupplots}
\usepgfplotslibrary{dateplot}
\usepgfplotslibrary{fillbetween}

\newcommand{\HT}{H\transpose}
\newcommand{\HHT}{HH\transpose}

\usepackage{acro}
\acsetup{single={1},	
	barriers/use = true,
	barriers/reset,
	barriers/single,
	uppercase/list}

\DeclareAcronym{RL}{short = RL, long = reinforcement learning, short-indefinite = an}
\DeclareAcronym{IQC}{short = IQC, long = integral quadratic constraint}
\DeclareAcronym{MPC}{short = MPC, long = model predictive control}
\DeclareAcronym{LQR}{short = LQR, long = linear quadratic regulator}
\DeclareAcronym{LTI}{short = LTI, long = linear time-invariant, short-indefinite = an}
\DeclareAcronym{BIBO}{short = BIBO, long = {bounded-input, bounded-output}}
\DeclareAcronym{SISO}{short = SISO, long = {single-input, single-output}}
\DeclareAcronym{MIMO}{short = MIMO, long = {multiple-input, multiple-output}}
\DeclareAcronym{SVD}{short = SVD, long = singular value decomposition}
\DeclareAcronym{PID}{short = PID, long = proportional-integral-derivative}
\DeclareAcronym{PI}{short = PI, long = proportional-integral}
\DeclareAcronym{YK}{short = YK, long = \YK}

\begin{document}


\begin{frontmatter}

\title{Stabilizing reinforcement learning control: A modular framework for optimizing over all stable behavior\tnoteref{label1}}
\tnotetext[label1]{Please cite the journal version in Automatica: \url{https://doi.org/10.1016/j.automatica.2024.111642}.}


%

\author[math]{Nathan P. Lawrence}
\ead{input@nplawrence.com}
\author[math]{Philip D. Loewen}
\ead{loew@math.ubc.ca}
\author[chbe]{Shuyuan Wang}
\author[honeywell]{Michael G. Forbes}
\author[chbe]{R. Bhushan Gopaluni}
\ead{bhushan.gopaluni@ubc.ca}

\address[math]{Department of Mathematics, University of British Columbia, Vancouver BC, Canada}
\address[chbe]{Department of Chemical and Biological Engineering, University of British Columbia, Vancouver, BC Canada}
\address[honeywell]{Honeywell Process Solutions, North Vancouver, BC Canada}

\begin{abstract}                          
We propose a framework for the design of feedback controllers that combines the
optimization-driven and model-free advantages of deep \acl{RL}
with the stability guarantees provided by using the \YK\ parameterization to define
the search domain.
Recent advances in behavioral systems allow us to construct a data-driven internal model;
this enables an alternative realization of the \YK\ parameterization based entirely on 
input-output exploration data.
Perhaps of independent interest, we formulate and analyze the stability of such data-driven models in the presence of noise.
The \YK\ approach requires a stable ``parameter'' for controller design.
For the training of \acl{RL} agents, the set of all stable linear operators is given explicitly through a matrix factorization approach.
Moreover, a nonlinear extension is given using a neural network to express a parameterized set of stable operators, which enables seamless integration with standard deep learning libraries.  
Finally, we show how these ideas can also be applied to tune fixed-structure controllers.
\end{abstract}
\end{frontmatter}
\acresetall
\section{Introduction}

\begin{figure}[tbp]
\centering
\includegraphics[width=8.4cm]{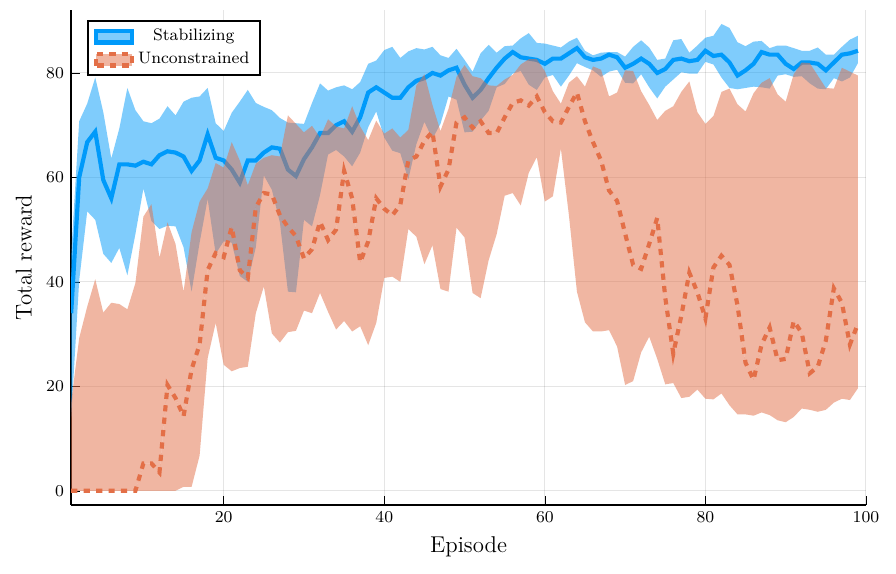}
\caption{
Cumulative reward over two PI-tuning experiments: one using the proposed stabilizing framework and the other using standard \acs{RL}. 
The stability-agnostic agent often destabilizes
the system and struggles to recover.
} 
\label{fig:rewards_PID}
\end{figure}

Closed-loop stability is a basic requirement in controller design. 
However, many learning-based control schemes do not address it explicitly \citep{busoniu2018ReinforcementLearning}.
This is somewhat understandable.
First, the ``model-free'' setup assumed in such algorithms, compounded by the complexity of the methods and their
underlying data structures, makes stability difficult to reason about.
Second, especially in the case of \ac{RL}, many of the striking recent success stories pertain
to simulated tasks or game-playing environments in which catastrophic failure has no real-world impact.
When a feedback controller is to be learned directly with \ac{RL}, 
system stability during exploration (along with learning performance) is influenced by the discount factor, reward function, and numerous other hyperparameters~\citep{busoniu2018ReinforcementLearning}.
\Cref{fig:rewards_PID} illustrates this point.
These issues provide a counterpoint to the generality and expressive capacity of modern \ac{RL} algorithms, which have nonetheless attracted immense interest for control tasks \citep{nian2020ReviewReinforcement}.
\par

\subsection{Contributions}

We propose a \emph{stability-preserving framework} for \ac{RL}-based controller design.
Our inspiration is the \ac{YK} parameterization \citep{anderson1998YoulaKucera}, which characterizes all stabilizing controllers for a given system.
We formulate a ``model-free'' realization of the \ac{YK} parameterization from exploration data, enabling \iac{RL} agent to optimize over all stable closed-loop behavior in an unconstrained fashion.
Specifically,
we leverage tools from the behavioral systems literature \citep{markovsky2021BehavioralSystems}: a Hankel matrix of input-output data serves as an internal model through a dynamic variation of Willems' fundamental lemma.
Under this regime, \iac{RL} agent is able to directly manipulate the closed-loop dynamics through a learnable stable operator.
We show how this stable operator can be deployed in an unconstrained and seamless fashion for both linear and nonlinear control strategies.

Perhaps of independent interest, we formulate a data-driven stability criterion in terms of the Hankel matrix structure commonly used in data-driven control. Output noise complicates the situation when working with Hankel matrices. 
We provide probabilistic analysis for the stability of such models as well. \par 

In sum, we disentangle three key components of \ac{RL}-based control system design: Algorithms, function approximators, and dynamic models.
Moreover, our framework supports a modular approach to learning stabilizing policies, in which advances in any single category can be applied to improve overall results. \par

\subsection{Related work}
\label{subsec:related}

\citet{busoniu2018ReinforcementLearning} provide a survey of \ac{RL} techniques from a control-theoretic perspective, emphasizing the need for stability-aware \ac{RL} algorithms.
Since one of the appeals of \ac{RL} is model-free policy optimization, methods for incorporating stability vary widely based on prior assumptions about the underlying dynamics.
As such, a wide variety of approaches have been proposed. 
Relatively early methods for incorporating stability into \ac{RL} are based on \acp{IQC} to capture nonlinearities or time-varying components in the environment or policy structure \citep{kretchmar2001RobustReinforcement, anderson2007RobustReinforcement}. 
In the context of \ac{RL}, nonlinearities in the environment or the nonlinear activation functions used to construct a policy neural network can be characterized using \acp{IQC}. 
This is also the basis for more recent approaches \citep{jin2020StabilitycertifiedReinforcement, revay2021RecurrentEquilibrium, gu2022RecurrentNeural, wang2022LearningAll}.
Lyapunov theory is another popular framework in the \ac{RL} literature \citep{huaguangzhang2011DataDrivenRobust, modares2014IntegralReinforcement, berkenkamp2017SafeModelbased, han2020ActorCriticReinforcement, kim2020ModelBased, chang2021StabilizingNeural, gros2022LearningMPC}.
The principal idea is to learn a policy that satisfies the decrease condition for a suitable Lyapunov function. 
Similarly, the \ac{LQR} is a fruitful testbed for benchmarking and analyzing \ac{RL} algorithms;
several works develop stability guarantees when the system dynamics are not available to the \ac{RL} agent \citep{perdomo2021StabilizingDynamical, lale2022ReinforcementLearning, mukherjee2022ReinforcementLearning}.\par

The \ac{YK} parameterization is seemingly an under-utilized technique for incorporating stability into \ac{RL} algorithms, with some examples due to \citet{roberts2011FeedbackController,friedrich2017RobustStability}.
\citet{roberts2011FeedbackController} propose its use after evaluating the performance of \ac{RL} with several different controller parameterizations for a simulated ball-catching task. 
Subsequently, \citet{friedrich2017RobustStability} employ the \ac{YK} parameterization through the use of a crude plant model; \ac{RL} is used to optimize the tracking performance of a physical two degrees of freedom robot in a safe fashion while accounting for unmodeled nonlinearities. 
Recently, a recurrent neural network architecture based on \acp{IQC} was developed \citep{revay2021RecurrentEquilibrium}. Since this architecture satisfies stability conditions by design, it can be used for control in a nonlinear version of the \ac{YK} parameterization \citep{wang2022LearningAll}.\par

While we also use the \ac{YK} parameterization, our approach has several novel aspects. 
We propose to produce stable operators using a non-recurrent neural network structure; this makes the implementation and integration with off-the-shelf \ac{RL} algorithms relatively straightforward, for both on-policy and off-policy learning. 
This contrasts with \ac{IQC} or Lyapunov-based approaches, such as \citet{jin2020StabilitycertifiedReinforcement, huaguangzhang2011DataDrivenRobust, modares2014IntegralReinforcement}, that place strong structural hypotheses on the 
network
architectures and update schemes.
We also formulate a data-driven realization of the \ac{YK} parameterization based on Willems' fundamental lemma, essentially removing the need for prior modeling, whereas the 
cited
works based on the \ac{YK} parameterization assume that a parameterized model is given. Moreover, we establish the stability of such data-based models, a principal hypothesis in the \ac{YK} parameterization. 
Finally, we show how the techniques presented here can be applied to a fixed-structure controller, an aspect not covered in other \ac{YK}-based approaches.

\subsection{Notation}

Given a matrix $M \in \reals^{m \times n}$,
we write $\norm{M}_{F} = \left(\sum_{i,j} m_{i,j}^2\right)^{\frac{1}{2}}$ for the Frobenius norm
and $\norm{M}$ for the spectral norm, that is, the largest singular value.
$M^{+}$ denotes the Moore-Penrose pseudoinverse. (Often $m \le n$ and $M$ has full rank, in which case $M^{+} = M^{\transpose} \left(M M^{\transpose}\right)^{-1}$.)
When $m=n$, we indicate the spectral radius by 
$\rho(M) = \max \left\{ \abs{\lambda} \colon \lambda \text{ is an eigenvalue of } M \right\}$. 
If $M=M\transpose$, we write $M>0$ (or $M\ge0$) instead of saying $M$ is positive-definite (resp. semi-definite).
\par

\section{Background}
\label{sec:background}

We consider a nominal \ac{LTI} system
whose state $x$ evolves in $\reals^n$:
\begin{equation}
P
\left\{\begin{aligned}	x_{t+1} &= A x_{t} + B u_{t} && \\
	y_{t} &= C x_{t}, && t=0,1,2,\ldots.
\label{eq:LTI}
\end{aligned}\right.
\end{equation}
The corresponding transfer function is
$P(z) = C (z I - A)^{-1} B$.
We treat the constant matrices $A,B,C$ as unknown,
and lay the foundation 
for Willems' fundamental lemma and the \ac{YK} parameterization
with the following mild assumptions.

\begin{assumption}\label{assump:unknown}
An upper bound of the state dimension $n$ is available.
\end{assumption}

\begin{assumption}\label{assump:contobs}
The matrix pair $(A, B)$ is controllable, and the pair $(A, C)$ is observable.
\end{assumption}

\begin{assumption}\label{assump:stablesiso}
The nominal system is stable and \ac{SISO},
that is, $\rho(A)<1$, $B\in\reals^{n\times 1}$, and $C\in\reals^{1\times n}$.
\end{assumption}

\subsection{A dynamic Willems' lemma as an internal model}

Given an $N$-element sequence $\{ z_{t} \}_{t = 0}^{N-1}$ of vectors in $\reals^m$
and an integer $L$, $1\le L\le N$, the \emph{Hankel matrix of order $L$} 
is the $mL \times (N - L + 1)$ array with the constant skew-diagonal structure
\[
H_{L}(z) =
\begin{bmatrix}
	z_{0} & z_{1} & \ldots & z_{N-L} \\
	z_{1} & z_{2} & \ldots & z_{N-L+1} \\
	\vdots & \vdots & \ddots & \vdots \\
	z_{L-1} & z_{L} & \ldots & z_{N-1}
\end{bmatrix}.
\label{eq:Hankel}
\]
Situations where this matrix has linearly independent rows are of particular interest.

\bdef\label{def:PE}
The sequence $\{ z_{t} \}_{t = 0}^{N-1} \subset \reals^{m}$ is \emph{persistently exciting of order $L$\/} if $\rank(H_L(z)) = mL$.
\edef 

\bdef
An input-output sequence $\{ u_{t}, y_{t} \}_{t = 0}^{N-1}$ is a \emph{trajectory} of \iac{LTI} system $(A, B, C)$ if there exists a state sequence $\{ x_{t} \}_{t = 0}^{N-1}$ such that \cref{eq:LTI} holds.
\label[definition]{def:trajectory}
\edef

The following theorem is the state--space version of Willems' fundamental lemma \citep{willems2005NotePersistency,markovsky2021BehavioralSystems}. 
It provides an alternative characterization of \iac{LTI} system based entirely on input-output data.

\bthm[See \citet{vanwaarde2020WillemsFundamental}]
Let $\{ u_{t}, y_{t} \}_{t = 0}^{N-1}$ be a trajectory of \iac{LTI} system $(A, B, C)$ where $u$ is persistently exciting of order $L+n$. Then $\{ \bar{u}_{t}, \bar{y}_{t} \}_{t = 0}^{L-1}$ is a trajectory of $(A, B, C)$ if and only if there exists $\alpha \in \reals^{N-L+1}$ such that
\[
\begin{bmatrix}
	H_L(u) \\
	H_L(y) 
\end{bmatrix}
\alpha =
\begin{bmatrix}
\bar{u} \\
\bar{y}	
\end{bmatrix}.
\label{eq:fundamentalLemma}
\]
\label{thm:fundamentalLemma}
Here the right-hand side is the block-structured column vector 
formed from $\bar u = \left[u_0 \ldots u_{L-1}\right]\transpose$ and $\bar y = \left[y_0 \ldots y_{L-1}\right]\transpose$.
\ethm

In applications, 
one uses measured input-output data to 
construct the left-hand side in \cref{eq:fundamentalLemma}.
Then, to test whether a candidate input-output sequence of length $L$ is indeed a system trajectory,
one uses it as the right-hand side and attempts to solve for $\alpha$ \citep{markovsky2021BehavioralSystems,vanwaarde2020WillemsFundamental,berberich2020TrajectorybasedFramework}.

We now formulate a dynamic variant of \cref{thm:fundamentalLemma}, enabling one to advance a trajectory in time.
Given $N$ and vectors $z_0,\ldots,z_N$, let 
$z = \{ z_{t} \}_{t = 0}^{N-1}$ and 
$z' = \{ z_{t} \}_{t = 1}^{N}$.
Then let
\[
H'_{L}(z) = H_{L}(z').
\]
Note that $H_L'(z)$ has the same shape as $H_L(z)$.

Given a system trajectory $\{ u_{t}, y_{t} \}_{t = 0}^{L-1}$ on the right-hand side of \cref{eq:fundamentalLemma} with $L \geq n$, we note that the next output $y_{L}$ is uniquely determined by these available data.
Intuitively, a time-shifted Hankel matrix advances the internal, unknown state of the system forward resulting in $y_{L}$.

\bcor
Let $\{ u_{t}, y_{t} \}_{t = 0}^{N}$ be a trajectory of a strictly proper \ac{LTI} system $(A, B, C)$ where $u$ is persistently exciting of order $L+1+n$. Then for each trajectory $\{ \bar{u}_{t}, \bar{y}_{t} \}_{t = 0}^{L-1}$ of $(A, B, C)$, there exists $\alpha \in \reals^{N-L+1}$ such that
\[
\bar{y}'
= H'_{L}(y)\alpha. 
\label{eq:proper}
\]
\label{cor:proper}
\ecor
\begin{remark}
    The hypotheses are to ensure both $\begin{bsmallmatrix} H_L(u) \\ H_L(y) \end{bsmallmatrix}$ and $\begin{bsmallmatrix}  H'_L(u) \\ H'_L(y) \end{bsmallmatrix}$ satisfy the requirements in \cref{thm:fundamentalLemma}. 
\end{remark}
\begin{proof}
Direct calculation following \cref{def:PE,thm:fundamentalLemma} as shown in \citet{lawrence2023ModularFramework}. Reproduced here for completeness.
By \cref{thm:fundamentalLemma}, the trajectory $\{ \bar{u}_{t}, \bar{y}_{t} \}_{t = 0}^{L-1}$ satisfies
\[
\begin{bmatrix}
	H_L(u) \\
	H_L(y) 
\end{bmatrix}
\alpha =
\begin{bmatrix}
\bar{u} \\
\bar{y}	
\end{bmatrix}
\]
for some $\alpha \in \reals^{N-L+1}$. Moreover, by \cref{def:trajectory} there exists a sequence of states $\{\bar{x}_{t}\}_{t=0}^{L-1}$ that corresponds to the input-output trajectory $\{ \bar{u}_{t}, \bar{y}_{t} \}_{t = 0}^{L-1}$. This sequence induces the state $\bar{x}_{L}$. We have
\begin{align}
	\sum_{i=0}^{N-L} \alpha_i y_{L+i} &=  \sum_{i=0}^{N-L} \alpha_i C \left( A x_{L-1+i} + B u_{L-1+i} \right) \\
	&= C \left( A \sum_{i=0}^{N-L} \alpha_{i} x_{L-1+i} + B \sum_{i=0}^{N-L} \alpha_{i} u_{L-1+i} \right) \\
	&= C\left( A \bar{x}_{L-1} + B \bar{u}_{L-1} \right) \\
	&= C \bar{x}_{L} \\
	&= \bar{y}_{L} 
\end{align}
as desired.
\end{proof}

\Cref{alg:sim} shows how to use this scheme in \cref{cor:proper} for closed-loop simulation. 
Moreover, this idea is particularly useful for aligning the true system with an internal Hankel representation. \par

{
\begin{minipage}[tbp]{\linewidth}
\linespread{1}
\normalsize
\SetArgSty{textnormal}
\begin{algorithm}[H]
\caption{Data-driven simulation}\label{alg:sim}
\SetKwInput{KwInput}{Input}
\KwInput{Data $\{u_k, y_k\}_{k=0}^{N}$ with persistently exciting input of order $L+1+n$; Initial trajectory $\{\bar{u}_k, \bar{y}_k\}_{k = 0}^{L-1}$}
\For{each time step}{
	Solve for $\alpha$: $\begin{bsmallmatrix}
		H_L(u) \\
		H_L(y) 
	\end{bsmallmatrix}
	\alpha =
	\begin{bsmallmatrix}
	\bar{u} \\
	\bar{y}	
	\end{bsmallmatrix}$\;
	Compute the next element $\bar{y}' = H'_{L}(y)\alpha$\;
	Generate the next control input $\bar{u}_{L}$\;
	Update trajectory: $\{\bar{u}_k, \bar{y}_k\}_{k = 0}^{L-1} \gets \{\bar{u}_k, \bar{y}_k\}_{k = 1}^{L}$\;
}
\end{algorithm}
\end{minipage}
}

\subsection{Data-driven realization of the Youla-Ku\v cera parameterization}
\label{subsec:dataYK}

We consider the standard four sensitivity functions associated with a plant $P$ and controller $K$: $\frac{P K}{1 + P K}, \frac{P}{1 + P K}, \frac{K}{1 + P K}, \frac{1}{1 + P K}$.
The \ac{YK} parameterization produces the set of all stabilizing controllers through a combination of an internal system model and a stable operator.
 The trick is to parameterize the aforementioned closed-loop transfer functions, then recover a controller. For example, the response of the transfer function $\frac{P K}{1 + P K}$ from the reference $r$ to output $y$ is determined by the transfer function $\frac{K}{1 + P K}$. By introducing a stable design variable $Q$, we can then directly shape the stable behavior of the system through the transfer function $PQ$. By asserting $Q = \frac{K}{1 + P K}$ and solving for $K$, we arrive at the \ac{YK}  parameterization \citep{anderson1998YoulaKucera}:
\[
\mathcal{K}_{\text{stable}} = \left\{ \frac{Q}{1 - Q P} \colon Q \text{ is stable} \right\}.
\label{eq:YK}
\]
Indeed, for a stable plant $P$, all four sensitivity functions are stable for any $K$ in $\mathcal{K}_{\text{stable}}$.
Moreover, when $P$ is linear, one may use a nonlinear operator $Q$ to parameterize nonlinear controllers \citep{anderson1998YoulaKucera,wang2022LearningAll}. \par

In \cref{alg:YK}, we translate the mathematical ideas above into a direct
sequential process.
In particular, we utilize \cref{cor:proper} in conjunction with the feedback connections in \cref{eq:YK} to produce stabilizing actions.
\Cref{thm:dataYK} provides details of the correspondence.

{
\begin{minipage}[]{\linewidth}
\linespread{1}
\normalsize
\SetArgSty{textnormal}
\begin{algorithm}[H]
\caption{Data-driven stabilizing controller}\label{alg:YK}
\SetKwInput{KwInput}{Input}
\KwInput{Stable parameter $Q$; Data $\{u_k, y_k\}_{k=0}^{N}$ with persistently exciting input of order $L+1+n$; Initial trajectory $\{\bar{u}_k, \bar{y}_k\}_{k = 0}^{L-1}$}  
\For{each time step $t$}{
	Set $u_{t-1} \gets \bar{u}_{L-1}$\;
	Observe the tracking error $e_t = r_t - y_{t}$ from the system\;\label{line:error}
	Compute $\bar{y}_L$ from \cref{eq:proper}\;\label{line:predict}
	Apply the input $\hat{r} = e_t + \bar{y}_{L}$ to the $Q$ parameter and return control action $\bar{u}_{L}$; for example, step forward in time of \iac{LTI} representation of $Q$\;\label{line:Qio} 
	Update the trajectory: $\{\bar{u}_k, \bar{y}_k\}_{k = 0}^{L-1} \gets \{\bar{u}_k, \bar{y}_k\}_{k = 1}^{L}$\;
}
\end{algorithm}
\end{minipage}
}
\begin{remark}
	Notice in \cref{line:Qio} that $Q$ ideally parameterizes the input-output dynamics between the reference $r$ and controls $u$. In practice, $Q$ takes into account discrepancies between the true plant output $y_t$ (\cref{line:error}) and the internal prediction $\bar{y}_L$ (\cref{line:predict}).
\end{remark}

\bthm
Assume $P$ is a stable and strictly proper \ac{LTI} system. Let $Q$ be a stable and proper \ac{LTI} parameter. Given an upper bound $L$ of the order of $P$, \cref{alg:YK} produces the same control signal $\{\bar{u}_t\}_{t=0}^{\infty}$ as the \ac{YK}  parameterization.
\label{thm:dataYK}
\ethm
\begin{proof}
We use $q_t$, $p_t$ to denote the impulse responses of $Q$ and $P$, respectively. Similarly, respective minimal state--space matrices are denoted $(A_{q}, B_{q}, C_{q}, D_{q})$ and $(A_{p}, B_{p}, C_{p})$.\par 
By the \ac{YK} parameterization, we have $U=KE$ for the controller $K \in \mathcal{K}_{\text{stable}}$ given by
\begin{align}
&& K(z) &= \frac{Q(z)}{1 - Q(z)P(z)}\quad \forall z \in \comps\nonumber \\
\iff&& \left(1 - Q(z)P(z) \right) U(z) &= Q(z) E(z)\nonumber \\
\iff&& u_t &= q_t * (e_t + p_t * u_t)\quad  \forall t \in \natsO\nonumber \\
&& &= \sum_{j=0}^{t-1} C_{q} A_{q}^{t-1-j} B_{q} \hat{r}_{j} + D_{q} \hat{r}_{t}, \label{eq:YKimpulse}
\end{align}
where $\hat{r}_{j} = e_{j} + \sum_{i=0}^{j-1} C_{p}A_{p}^{j-1-i}B_{p} u_i$ and $*$ is the convolution operator; we have also assumed, without loss of generality, that $P$ and $Q$ have zero initial state.\par
Next we relate \cref{eq:YKimpulse} to \cref{alg:YK}.
Let $\{ e_k \}_{k=0}^{\infty}$ be an arbitrary sequence. (Such a sequence is dynamically generated in \cref{alg:YK}.)
Without loss of generality, let the initial trajectory be $\{\bar{u}_k, \bar{y}_k\}_{k = 0}^{L-1} = \{0, 0\}_{k = 0}^{L-1}$. For each time $t \in \natsO$ we compute $\alpha^{(t)}$ and $\bar{y}_{t} = \bar{y}_{L}$ from \cref{eq:proper}. 
Since $L$ is an upper bound on the order of $P$, 
$\bar{y}_{t}$ is the unique next output from the trajectory $\{\bar{u}_k, \bar{y}_k\}_{k = 0}^{L-1}$. 
Therefore, we have $
	\hat{r}_{t} = e_{t} + \sum_{i=0}^{N-L} \alpha_{i}^{(t)} y_{L+i}.$
Then $\bar{u}_t = \sum_{j=0}^{t-1} C_{q} A_{q}^{t-1-j} B_{q} \hat{r}_{j} + D_{q} \hat{r}_{t}$ gives the next control input. \par
By updating the trajectory between time steps---$\{\bar{u}_k, \bar{y}_k\}_{k = 0}^{L-1} \gets \{\bar{u}_k, \bar{y}_k\}_{k = 1}^{L}$---we dynamically generate a sequence $\{ \alpha^{(t)} \}_{t=0}^{\infty}$ that produces the control inputs $\{ \bar{u}_{t} \}_{t=0}^{\infty}$ satisfying the discrete integral equation in \cref{eq:YKimpulse}.
\end{proof}

\section{On the stability of noisy Hankel matrices}
\label{sec:stableHankel}

\begin{figure}[tbp]
\centering
\includegraphics[width=8.4cm]{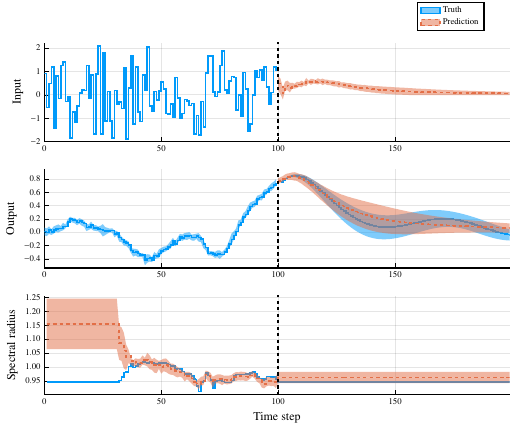}
\caption{
$100$ time steps of input-output data are collected using a standard normal probing signal.
The recursion in \cref{eq:alpha_dynamics} is used to continue the rollout. This is done several times for different samples of output noise.
The bottom figure is the evolution of the spectral radii for the noisy and noise-free matrices $H^{+} H'$.} \label{fig:stableRollout}
\end{figure}

\Cref{thm:dataYK} assumes the underlying system is open-loop stable, in which case one may utilize \cref{alg:YK} to produce stabilizing control actions. 
However, the long-term predictions generated by \cref{cor:proper} will be influenced by the noise in the data and singular values of the resulting stacked Hankel matrices.
Stopping the data collection process early can result in unstable predictions even for an open-loop stable plant; see the initial spectral radius values in \cref{fig:stableRollout}.

\subsection{Data-driven stability test}
\label{subsec:stabletest}
We formulate the stability of a Hankel matrix system representation by delving deeper into the recursive nature of infinite trajectories generated by \cref{alg:sim}. 
This involves deriving a special matrix structure that relates successive solutions $\alpha_{t}, \alpha_{t+1}, \ldots$ from \cref{eq:fundamentalLemma}. \par

Start with the equation from \cref{thm:fundamentalLemma}: 
let $\alpha_0$ denote the minimum-norm solution of
\[
\begin{bmatrix} H_L(u) \\ H_L(y) \end{bmatrix} \alpha_{0} 
= \begin{bmatrix} \bar{u} \\ \bar{y} \end{bmatrix}.
\]
(Henceforth we assume minimum-norm solutions; any solution may be used, but the minimum-norm solution will lead to a clean formulation.)
By \cref{cor:proper} we then have that the successive output trajectory is given by $\bar{y}' = H'(y) \alpha_{0}.$ By extension, the next trajectory is given by $\begin{bsmallmatrix} \bar{u}' \\ \bar{y}' \end{bsmallmatrix} = \begin{bsmallmatrix} H'_L(u) \\ H'_L(y) \end{bsmallmatrix} \alpha_{0}$.

Starting from $\begin{bsmallmatrix} \bar{u}' \\ \bar{y}' \end{bsmallmatrix}$, we repeat the process to arrive at the recursion
\[
\underbrace{
\begin{bmatrix}
	H_L(u) \\
	H_L(y) 
\end{bmatrix}}_{H}
\alpha_{t+1} =
\underbrace{
\begin{bmatrix}
	H'_{L}(u) \\
	H'_{L}(y) 
\end{bmatrix}}_{H^{'}}
\alpha_{t},
\qquad t=0,1,2,\ldots.
\label{eq:alpha_dynamics}
\]
This can be seen as the ``free response'' of the $\alpha_{t}$ dynamics inferred from the collected data and produced by \cref{alg:sim}.\footnote{To evaluate the free response of the dynamics in \cref{eq:LTI} in Hankel form, one may add the constraint $H'_{L}(u) \alpha_{t} = 0$ to \cref{eq:alpha_dynamics}.}
Therefore, by checking the eigenvalues of $H^{+} H'$ we determine if the matrix transformation from a system's Hankel matrix to its time-shifted counterpart is internally ``contractive''; under the assumption of minimum-norm solutions, this implies that the behavior $\begin{bsmallmatrix} \bar{u} \\ \bar{y} \end{bsmallmatrix}$ is bounded and $\begin{bsmallmatrix} \bar{u} \\ \bar{y} \end{bsmallmatrix} \to 0$ as $t \to \infty$.
In the ensuing sections, we formalize and prove properties about this special matrix structure in the presence of noise. \par

\subsection{Random Hankel matrices}

Randomness complicates the notion of stability.
Going forward, we assume the outputs have the form $y_t + \omega_t$ where $\omega_t$ is normally distributed.
In order to characterize the eigenvalues of $H^{+} H'$ under measurement noise, we first isolate the underlying random Hankel matrix in the term $H_L(y) + H_L(\omega)$.
We will then be able to relate properties of the random matrix $H_L(\omega)$ to the overall structure.
Therefore, this subsection focuses on Hankel matrices of purely random signals as $N \to \infty$, and then the latter section re-introduces the input-output dynamics. \par

Given a sequence of independent random variables 
$\omega_0,\omega_1,\ldots$,
we consider the doubly-infinite array
\[
\mathcal{H} =
\bbmat
\omega_0 & \omega_1 & \omega_2 & \cdots \\
\omega_1 & \omega_2 & \omega_3 & \cdots \\
\omega_2 & \omega_3 & \omega_4 & \cdots \\
\vdots   &  \vdots  &  \vdots  & \ddots \\
\ebmat.
\label{eq:infiniteH}
\]
Our focus is on short wide submatrices anchored
at the top left corner.
Specifically, 
for any fixed positive integers $L$ and $N$,
we write $H_{L,N}$ for the top left $L\times N$ 
submatrix of $\mathcal{H}$.

\Cref{lem:HW} is a fundamental concentration inequality that drives our analysis of random Hankel matrices. \Cref{cor:HW} is a useful special case.

\blem[Hanson--Wright inequality, adapted from \citep{rudelson2013HansonWrightInequality}]
There exists a constant $c>0$ such that, 
for every $n\times n$ matrix $M$,
any random vector $X = (X_0, \ldots, X_{n-1}) \in \reals^{n}$ with independent standard normal components $X_i$
obeys
\[
\Pr\set{ \abs{X\transpose  M X - \mathbb{E} \left[ X\transpose  M X \right]} > t } \leq 2 \exp\left( -c \min \left\lbrace \frac{t^2}{\norm{M}_{F}^{2}}, \frac{t}{\norm{M}} \right\rbrace \right),\qquad t\ge0.
\label{lem:HW}\]
\elem

\bcor
Let $X_0,X_1,\ldots$ be a sequence of standard normal random variables.
Then there exist constants $c_0,c_1>0$ such that for any $n\in\nats$
and any $\alpha\in(0,1)$, one has both
\begin{enumerate}
\item[\textrm{(a)}]
$\displaystyle \Pr\set{ \abs{\sum_{k=0}^{n-1} X_k^2 - n} < \alpha n }
\ge 1 - 2\exp\left(-c_0\alpha^2 n\right)$, 
and
\item[\textrm{(b)}]
$\displaystyle \Pr\set{ \abs{\sum_{k=0}^{n-1} X_k X_{\sigma(k)}} < \alpha n }
\ge 1 - 2\exp\left(-c_1\alpha^2 n\right)$,
\\
for any $\sigma\colon\nats\to\nats$ such that $\sigma(k)\ne k$ for all $k$.
\end{enumerate}
\label{cor:HW}
\ecor

\begin{proof}
Both parts follow from taking $t=\alpha n$ in \cref{lem:HW}, 
and using $0<\alpha<1$ to simplify $\min\left\{ \alpha^2 n, \alpha n\right\} = \alpha^2 n$.
In part (a), one uses the $n\times n$ identity matrix for $M$:
clearly $\norm{I}_F^2 = n$ and $\norm{I}=1$.
In part (b), one defines $K=\max\{\sigma(1),\ldots,\sigma(n)\}$
and forms $M$ as a $K\times K$ matrix in which every entry is $0$
except for the $n$ entries at positions $(k,\sigma(k))$,
each of which equals $1$. 
Again $\norm{M}_F^2 = n$ and $\norm{M}=1$.
\end{proof}

\Cref{prop:minsigma} analyzes the limiting behavior of the singular values of random Hankel matrices as the number of samples tends to infinity.
Its proof, shown in \cref{app:randomHankel}, is a key step
toward a spectral analysis of an
interesting combination of related matrices.
\begin{restatable}[]{proposition}{minsigma}
Suppose each of the independent random variables $\omega_i$ has a standard normal distribution.
Then for each fixed $L\ge 1$, 
there is a sequence $r_0,r_1,\ldots$,
with $r_N\to\infty$ as $N\to\infty$, such that
\[
\lim_{N\to\infty} \Pr\set{ \sigma_{\rm min}\left(H_{L,N}\right) > r_N } 
= 1.
\]
Here $\sigma_{\rm min}(\cdot)$ returns the smallest singular
value of its matrix argument.
\label{prop:minsigma}	
\end{restatable}

Recall the matrix $\mathcal{H}$ in \cref{eq:infiniteH}; for any fixed positive integers $L$ and $N$,
we consider the top left submatrix of shape $L\times(N+1)$,
namely, $\bar{H}_{L,N} = H_{L, N+1}$,
and extract two $L\times N$ chunks of interest.
These are the ``standard'' Hankel matrix $H = H_{L,N}$
discussed above,
formed by removing the last column,
and the ``time-shifted'' Hankel matrix $H' = H'_{L,N}$,
formed by dropping the first column instead.
We are now ready to tackle the special matrix structure discussed in \cref{subsec:stabletest}, specifically, the eigenvalues of 
the random matrix $H^+ H' = \HT (H \HT )^{-1} H'$ for large $N$.

These constructions can be expressed as matrix products
as follows: 
\[
H  = \bar H W,\quad
H' = \bar H W',\qquad
\text{where}\quad
W  = \bbmat I \\ 0 \ebmat,\quad
W' = \bbmat 0 \\ I \ebmat.
\label{eq:extractHL}\]

Since cyclic permutations of matrix products preserve the spectral radius, we have
$$
\rho\left( H^{+} H' \right) 
= \rho\left( W' W\transpose  \bar{H}\transpose  \left( \HHT  \right)^{-1} \bar{H} \right).
$$
We estimate the right side using the general matrix relations
$$
\rho(M) 
\le \norm{M} 
= \sigma_{\max}(M) 
= \sqrt{\rho(M\transpose M)}.
$$
Thus we have
\[
\rho\left(H^+ H'\right)^2 
\le \rho\left(\bar{H}\transpose  \left( \HHT  \right)^{-1} \bar{H}\right).
\label{eq:spectralupperbound}
\]
We recognize the inverse matrix filling the sandwich on the right side
from \cref{prop:minsigma}; the symmetry and nested structure in \cref{eq:spectralupperbound} supports the following result.
(See \cref{app:randomHankel} for the proof.)

\begin{restatable}[]{theorem}{HplusHprime}
Suppose each of the independent random variables $\omega_i$ has a standard normal distribution.
Then for each fixed $L\ge 1$, 
there is a sequence $\eps_0,\eps_1,\ldots$,
with $\eps_N\to0$ as $N\to\infty$, such that
\[
\lim_{N\to\infty} \Pr\set{ \rho\left(H_{L,N}^+ H'_{L,N}\right) < 1 + \eps_N } 
= 1.
\]
\label{prop:HplusHprime}
\end{restatable}

\subsection{Hankel models with additive noise}

We are now ready to re-introduce the system dynamics and establish the stability of the resulting Hankel-based models following \cref{eq:alpha_dynamics}. 
To provide intuition for the concept, \cref{fig:stableRollout} illustrates \cref{cor:noiseHankel} and \cref{prop:HplusHprime} by visualizing the recursion in \cref{eq:alpha_dynamics}.

\bcor[Stable Hankel dynamics models]
In addition to \cref{assump:unknown,assump:contobs,assump:stablesiso}, take $L \geq n$ and assume the outputs $y_{t}$ have additive Gaussian noise $\omega_{t}$. Assume the probing signal $u$ is bounded for all time and satisfies:
\[
\lambda_{\min}\left( H_{L}(u) H_{L}(u)\transpose \right) \to \infty.
\label{eq:probeCond}
\]
Denote $H$ to be the principal matrix in \cref{eq:fundamentalLemma}. Then the result of \cref{prop:HplusHprime} still holds.
\label{cor:noiseHankel}
\ecor
\begin{remark}
A truncated Gaussian probing signal satisfies the hypotheses by \cref{prop:minsigma}.
\end{remark}
\begin{proof}
Let $z'$ be the last column of $H'$:
\[
z' = \left[ u_{N-L} \ldots u_{N-1}, y_{N-L} + \omega_{N-L} \ldots y_{N-1} + \omega_{N-1} \right]\transpose.
\]
Using the relation $H^+ = \lim_{\delta \downarrow 0} \HT \left( H\HT + \delta I \right)^{-1}$ to modify \cref{eq:stablewall} in \cref{prop:HplusHprime}, we arrive at the general inequality
\[
\rho(H^+H')
\le
\sigma_{\max}\left( H^+ H' \right) 
\leq \sqrt{1 + \norm{ z' }^2\norm{H^+}^2}.
\label{eq:stablewall_repeat}
\]
We are interested in the top $L+n$ singular values of $H$.
Following \cref{prop:minsigma} and \citet{coulson2022quantitative}, it can be shown that $\sigma_{L+n}(H) \to \infty$, meaning $\norm{H^+} \to 0$.
However, this is insufficient for \cref{eq:stablewall_repeat} to converge to $1$. Since the system of interest is \ac{BIBO} stable, we have
\[
\norm{z'}^{2} \leq 2L C + \norm{\omega'}^2,
\]
where $C$ is a constant and $\omega'$ is the $L$-dimensional vector of noise terms in $z'$. The result then follows.
\end{proof}
\begin{remark}
If the system is unstable, then we expect an exponential increase in the magnitude of $z'$, blowing up our spectral radius estimate in \cref{eq:stablewall_repeat}.
\end{remark}

\section{Stabilizing reinforcement learning control}
\label{sec:stableRL}

The \ac{YK} parameterization in \cref{eq:YK} features two ingredients for the set of stabilizing controllers: the dynamics $P$ and the stable operator $Q$.
The previous two sections showed how to incorporate Willems' lemma to characterize $P$.
This leaves $Q$ as the ``learnable'' component for \iac{RL} agent.
The advantage of learning $Q$ over a standard feedback policy is it enables \iac{RL} agent to update its policy in an unconstrained fashion without risking instability during training.

\subsection{Learning stable operators}
\label{sec:Qparam}

The $Q$ parameter is a dynamical system.
Therefore, $Q$ is characterized by inputs, outputs, and some stable internal transition.
We demonstrate two approaches for modeling stable internal dynamics amenable to deep learning and optimization frameworks: one for the linear case, then an extension to the nonlinear setting.
In both cases, we make use of Lyapunov's second method: the main idea is to embed a trainable Lyapunov function inside the dynamic model.

Let us recall the definition of a \emph{Lyapunov candidate function} $V\colon \reals^n \to \reals$: 1) $V$ is continuous; 2) $V(z) > 0$ for all $z \neq 0$, and $V(0) = 0$; 3) There exists a continuous, strictly increasing function $\varphi\colon[0, \infty) \to [0, \infty)$ such that $V(z) \geq \varphi(\norm{z})$ for all $z \in \reals^n$; 4) $V(z) \to \infty$ as $\norm{z} \to \infty$.
\par

\emph{(Linear operators)}\quad
We consider stable linear operators of the form
\begin{equation}
Q
\left\{\begin{aligned}
	z_{t+1} &= A_q z_{t} + B_q (e_t + \bar{y}_L) \\
	u_{t} &= C_q z_{t} + D_q (e_t + \bar{y}_L)
\label{eq:QLTI}
\end{aligned}\right.
\end{equation}
where $\bar{y}_L$ is the latest internal prediction, for example, from \cref{alg:YK}.
Therefore, the parameterization of $Q$ is tied to the representation of stable matrices $A_q$.
However, the explicit representation of stable matrices is unwieldy:
${\mathcal{S}_{n} = \{ A_q \in \reals^{n \times n} \colon \rho(A_q) < 1 \}}$.
Indeed, $\mathcal{S}_{n}$ is non-convex and neither open nor closed.

Fix an arbitrary square matrix $\hat{M} \in \reals^{n \times n}$ and a lower triangular matrix $L \in \reals^{n \times n}$ with positive diagonal entries.
Consider the transformation ${M \leftarrow U \texttt{tanh}(D) V\transpose}$, based on the \ac{SVD} ${\hat{M} = U D V\transpose}$, where $\texttt{tanh}$ is applied componentwise.
Then the matrix $A_q = L^{-1} M L$ directly parameterizes the Lyapunov decrease condition ${A_{q} L^{-1}L\invtranspose A_{q}\transpose - L^{-1}L\invtranspose < 0}$ under the quadratic function ${V(z) = z\transpose L^{-1}L\invtranspose z}$.
Therefore, we have the following result:
\begin{align}
\begin{split}
\mathcal{S}_{n} = \{ L^{-1} U D V\transpose L \in \reals^{n \times n} \colon L > 0 \text{ lower triangular},\\ U \text{ and } V \text{ orthogonal},\ D \text{ diagonal and } \norm{D} < 1 \}.
\label{eq:stableMat_svd}
\end{split}
\end{align}
This is a corollary based on \citet{gillis2019ApproximatingNearest}.

\emph{(Nonlinear operators)}\quad For the problem of learning stable nonlinear operators, we adapt the method of \citet{lawrence2020AlmostSurely}: the idea is to construct stable autonomous systems of the form $z_{t+1} = f_{\theta}(z_t)$ ``by design'' through the use of trainable Lyapunov functions. ($\theta$ represents a set of trainable weights.)

In the present setup, $f_{\theta}$ models the internal dynamics of a nonlinear $Q$ parameter. 
For example, a control-affine model may be used with stable transition dynamics $f_{\theta}$ \citep{sontag1989smooth}.
The interpretation of a nonlinear $Q$ parameter is the same as the original motivation for $\mathcal{K}_{\text{stable}}$ in \cref{eq:YK}. 
The underlying interconnections remain the same, except now $Q$ characterizes nonlinear controllers.

Two neural networks work in tandem to form a single model that satisfies the decrease condition central to Lyapunov's second method: a smooth neural network $\hat{f}_{\theta}$, and a convex Lyapunov neural network $V_{\theta}$. Set $\hat{z}' = \hat{f}_{\theta}(z)$ where $z$ is the current ``state'' and $\hat{z}'$ is the proposed next state. Two cases are possible: either $\hat{z}'$ decreased the value of $V$ or it did not. We can write out a correction to the dynamics in closed form by exploiting the convexity of $V$:
\begin{align}
\begin{split}
    z_{t+1} &= f_{\theta}( z_t )\\
    & \equiv
    \begin{cases}
    \hat{f}_{\theta}(z_t), &\text{if } V( \hat{f}_{\theta}(z_t) ) \leq \beta V( z_t )\\
    \hat{f}_{\theta}(z_t)\left( \frac{\beta V( z_t )}{V( \hat{f}_{\theta}(z_t) )} \right), &\text{otherwise}
    \end{cases}\\
     &= \gamma \hat{f}_{\theta}( z_t ), \text{ where}\\\
     & \gamma=\gamma( z_t )= \frac{\beta V( z_t ) - \texttt{ReLU}\left( \beta V( z_t ) - V( \hat{f}_{\theta}( z_t ) ) \right)}{ V( \hat{f}_{\theta}( z_t ))}.
\label{eq:det_stable_scale}
\end{split}
\end{align}
(Recall $\texttt{ReLU}(x) = \max \{0,x \}$.)
Since \cref{eq:det_stable_scale} composes the model $f_{\theta}$, both $\hat{f}_{\theta}$ and $V_{\theta}$ are trained in unison towards whatever goal is required of the sequential states $z_t, z_{t+1}, \ldots$, such as supervised learning tasks. Moreover, although the model $f_{\theta}$ is constrained to be stable, it is unconstrained in parameter space, making its implementation and training fairly straightforward with deep learning libraries.

\subsection{Unconstrained reinforcement learning over stable operators}
\label{subsec:RL}

The \ac{YK} parameterization is appealing for learning-based control schemes such as \ac{RL} because the closed-loop system is stable for every choice of the $Q$ parameter. Therefore, stability does not rely on hyperparameter selection or optimality. This is in contrast to simply selecting a feedback controller without enforcing stability; see \cref{fig:rewards_PID}. Since any practical objective will require closed-loop stability, it is reasonable to allow \iac{RL} agent to manipulate the $Q$ parameter directly.
A brief overview of deep \ac{RL} will serve to unify this paper, however, a thorough introduction is beyond its scope.\par

\Ac{RL} is an optimization-driven framework for learning ``policies'' simply through interactions with an environment \citep{busoniu2018ReinforcementLearning,nian2020ReviewReinforcement}. The states $s$ and actions $a$ belong to the state and action sets $\mathcal{S}$, $\mathcal{A}$, respectively.
At each time step $t$, the state $s_t$ influences the sampling of an action $a_t \sim \pi(\cdot \mid s_t)$ from the ``policy'' $\pi$. Given the action $a_t$, the environment produces a successor state $s_{t+1}$, which induces a conditional density function $s_{t+1} \sim p( \cdot \mid s_t, a_t )$ for any initial distribution $s_{0} \sim p_{0}( \cdot )$.
The desirability of a given action is quantified by a ``reward'' $r_t = r(s_t, a_t)$ associated
with each step in the process above.
This cycle produces one step in a Markov decision process. As time marches forward under a policy $\pi$, a ``rollout'' emerges, denoted $h = (s_0, a_0, r_0, s_1, a_1, r_1, \ldots )$. Each fixed policy $\pi$ induces a probability density $p^\pi(\cdot)$ on the set of rollouts.\par

With these pieces in place, the overall goal of the agent is to determine a policy that maximizes the cumulative discounted reward.
That is, given some constant $\gamma\in(0,1)$, 
the agent seeks $\pi$ to 
\begin{equation}
\begin{aligned}
    &\text{maximize} && J(\pi) = \mathbb{E}_{h \sim p^{\pi}}\left[ \sum_{t=0}^{\infty} \gamma^{t}r(s_t,a_t) \right]\\
    &\text{over all} && \text{policies } \pi \colon \mathcal{S} \to \mathcal{P}(\mathcal{A}),
\end{aligned}
\label{eq:RLobjective}
\end{equation}
where $\mathcal{P}(\mathcal{A})$ denotes the set of probability measures on $\mathcal{A}$.\par

In the space of all possible policies, the optimization is performed over a subset parameterized by some vector $\theta$. 
In this work, the policy is the $Q$ parameter outlined in \cref{sec:Qparam}. 
Therefore, \cref{eq:RLobjective} automatically satisfies an internal stability constraint over the whole weight space $\theta$. 
We are then able to use any \ac{RL} algorithm to solve the problem. \par 

The broad subject of \ac{RL} concerns iterative methods for choosing a desirable policy $\pi$ (this is the ``learning''), guided in some fundamental way by the agent's observations of the rewards from past state-action pairs (this provides the ``reinforcement'').
A standard approach to solving Problem~\eqref{eq:RLobjective} uses gradient ascent
\[
\theta
\leftarrow \theta + \eta\nabla J(\theta),
\label{eq:PolicyGradient_Iteration}
\]
where $\eta > 0$ is a step-size parameter.
Analytic expressions for $\nabla J(\theta)$ exist for both stochastic and deterministic policies \citep{silver2014DeterministicPolicy}. 
However, $\nabla J(\theta)$ cannot be evaluated precisely, as it depends on the dynamics, policy, and chosen time horizon, not to mention the noise.
Therefore, \ac{RL} algorithms differ based on how they approximate the update scheme in \cref{eq:PolicyGradient_Iteration}. \par

Since our framework decouples stability from the learning process, one may employ any off-the-shelf \ac{RL} algorithm. 
Therefore, as the field of deep \ac{RL} matures, this stabilizing framework will remain relevant. 
The only requirement is an appropriate policy representation. 
Both the linear and nonlinear cases discussed in \cref{sec:Qparam} can be implemented in a standard \ac{RL} library: one must store the internal state $z_t$ and input $e_t + \bar{y}_L$, then employ automatic differentiation to update the $A_q, B_q, C_q, D_q$ matrices in \cref{eq:QLTI} or, in the nonlinear case, $f_\theta$ in a control-affine setup. Further details are provided in \cref{app:implementation}.

\section{Simulation studies}
\label{sec:results}

We now demonstrate the proposed stabilizing framework in a series of simulation studies.
We give an industrial example, showing how one can layer the stabilizing strategy on top of existing controllers.
Then, we show how the ideas presented above can be adapted to directly modify fixed-structure controllers while ensuring stability.
In all the examples, we use the TD3 algorithm \citep{fujimoto2018AddressingFunction}.
This choice is primarily to illustrate the applicability of the framework to general algorithms.
Note that the choice of \ac{RL} algorithm is essentially a hyperparameter layered on top of the stable behavior it modifies.
Code is available here: \url{https://github.com/NPLawrence/StableBehavior.jl}


\subsection{An industrial example}
\label{subsec:extank}

\begin{figure}[tbp]
\begin{center}
\includegraphics[width=8.4cm]{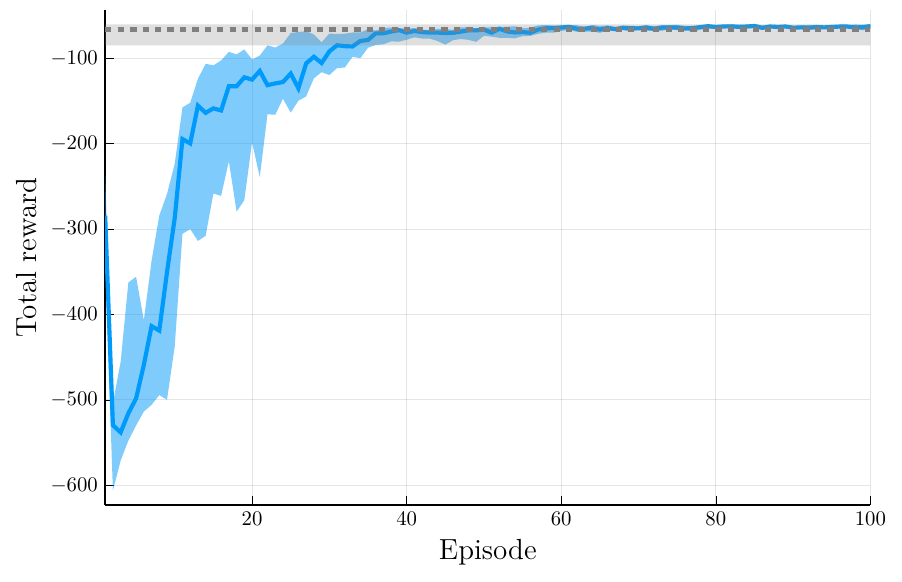}    
\caption{Cumulative reward curve over $20$ training sessions. The solid line is the median and the shaded region shows the interquartile range. The dashed line and its shaded region are the final results of training without the stability constraint.} 
\label{fig:rewards_tank}
\end{center}
\end{figure}

The authors' industrial partner built a hardware platform to use for testing various control methods.
The equipment involves a tank holding water,
positioned above a second tank used as a reservoir.
Water drains from the upper tank into 
the reservoir through an outflow pipe,
while being replenished by water pumped up 
from the reservoir.
The problem is to control the water level in the upper tank.

Two \ac{PID} controllers are in operation.
First, a ``level controller'' measures the 
actual water level outputs the desired inflow rate.
Second, a ``flow controller'' uses the desired
and actual inflow rates to determine the pump speed.
For our purposes, both these controllers are fixed and a part of the environment.

We have reliable numerical models for all
aspects of the equipment described above.
The flow dynamics, 
based on Bernoulli's equation and conservation of fluid,
are nonlinear.
Low-pass filtering leads to a stream of four
scalar signals: the water level, drainage flow rate, pump speed, and incoming flow rate.
A full account of the apparatus and the
differential equations we use to model it
appears in \cite{lawrence2022DeepReinforcement}.
For the results presented here, we used the
simulator rather than the laboratory system.
This involved discretizing the continuous dynamics 
cited above with time steps of $0.5$ seconds
and adding Gaussian measurement noise with variance $0.015$.
\par


We use the proposed stabilizing framework
to generate additive corrections to the
command produced by the given level \ac{PID} controller.
Since the environment includes \iac{PID} controller, we modify the control scheme to be in incremental form $u_t = u_{t-1} + \Delta u_t$, where $\Delta u_t$ is the sum of the nonlinear \ac{YK}  parameter from \cref{sec:Qparam} and \ac{PID} controller outputs:
\[
\Delta u_t = \Delta u^{(q)}_t + \Delta u^{(\text{PID})}_t
\]

\begin{figure}[tbp]
\begin{center}
\includegraphics[width=8.4cm]{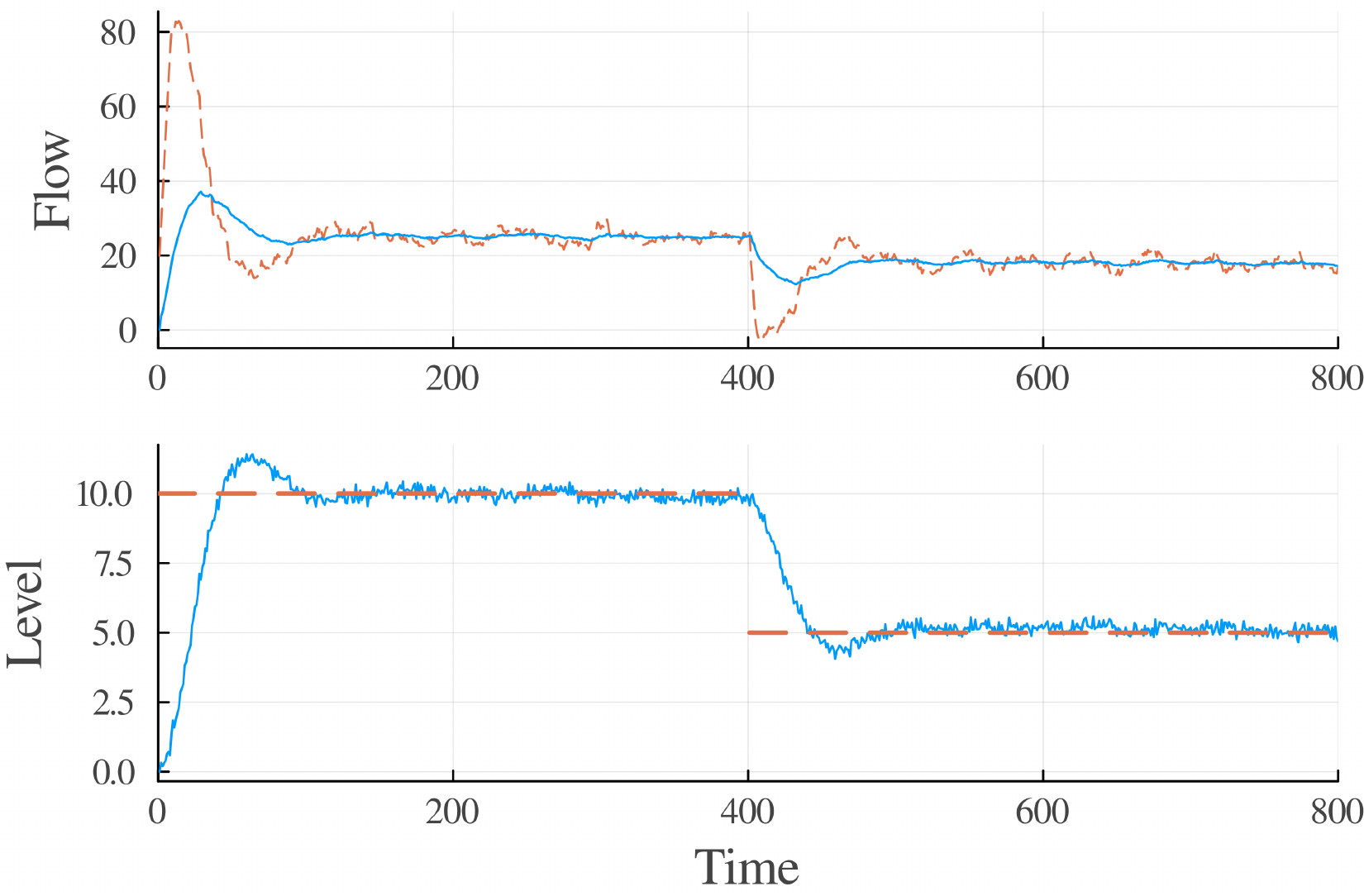}    
\caption{A sample input-output rollout by the trained \ac{RL} agent for one of the training sessions. Dashed lines are setpoints; solid lines are measured values.} 
\label{fig:input_output}
\end{center}
\end{figure}

Although the control system contains several cascaded filter terms,
the full flow setpoint to measured level dynamics is approximately a first-order plus dead time system \citep{lawrence2022DeepReinforcement}.
Recall Willems' lemma only requires an upper bound of the system order.
We take $L=11$ to ensure input--output trajectories are sufficiently long to capture the current dynamics in the presence of output noise.
We ran $20$ training sessions, each of $100$ episodes.
\Cref{fig:rewards_tank} illustrates the cumulative rewards observed. The median over the $20$ sessions provides the solid line; the interquartile ranges delimit the shaded region.
We note that the median reward curve is much closer to the upper limit of the shaded region than the lower, indicating that the majority of experiments fall within that tight region. Although there is significant change in the first few episodes, due to the random policy initialization, the training sessions exhibit consistent convergence. The reward curves tend to plateau after around $40$ episodes. \Cref{fig:input_output} shows a single rollout from one of the experiments. 

\subsection{Direct tuning of fixed-structure controllers}
\label{subsec:exPIDtune}

In the introduction, we highlighted the potentially dangerous dependence of closed-loop stability on hyperparameter settings. 
So far we have decoupled stability and learning algorithms through a data-driven control scheme. 
However, one may wish to enforce a fixed-structure control law. We show how our framework can also deal with this case through a data-driven constraint.
\bthm[\Acs{SISO} case of \citet{furieri2019InputOutputParametrization}]
Consider the set of
scalar-valued transfer functions $X, Y, W$ satisfying the linear relation
\begin{equation}
\begin{aligned}
	X + PY &= I \\
	W - PX &= 0.
\end{aligned}	
\label{eq:affinesub}
\end{equation}
Then
\[
\mathcal{K}_{\text{stable}} = \left\{ YX^{-1} \colon \cref{eq:affinesub} \text{ holds and } X, Y, W \text{ are stable} \right\}.
\label{eq:affineYK}
\]
\label{thm:furieri}
\ethm
\begin{remark}
By identifying $X = \frac{1}{1 + P K}, Y = \frac{K}{1 + P K}	, W = \frac{P}{1 + P K}$, we see that \cref{eq:affineYK} implicitly parameterizes all stable sensitivity functions in \cref{subsec:dataYK}.
\end{remark}

In contrast to the \ac{YK} parameterization in \cref{eq:YK}, \cref{thm:furieri} characterizes the set of stabilizing controllers through the affine constraint in \cref{eq:affinesub}. This alternative representation is useful for imposing a desired controller structure through the variables $X, Y$ while enforcing closed-loop stability by insisting $X, Y, W$ be stable. We use the linear parameterization from \cref{eq:stableMat_svd}.
\par

In the behavioral setting, we propose to traverse \cref{eq:affineYK} through the use of \cref{alg:sim} and the set of stable parameters in \cref{eq:QLTI}.
Concretely, we generate the left-hand side of \cref{eq:affinesub} by taking the outputs of $X, Y, W$ as inputs to the Hankel-based model in \cref{alg:sim}. We minimize the residual from the right-hand side to generate a stabilizing controller.
Note this approach can be used to find an initial stabilizing controller, to be deployed in combination with the control scheme shown in \cref{subsec:extank}.

\emph{(Training)}\quad
Consider a plant whose continuous-time transfer function is
\[
P(s) = \dfrac{1-s}{(s+1)^3}.
\]
Like the previous example, we discretize in time and take the resulting system as the true dynamics.
\par

We illustrate the data-driven stability constraint on \iac{PI} tuning task. 
The deterministic policy has the form
\[
\pi_{\text{PI},k_p,k_i}(s_t) 
= k_p (e_t - e_{t-1}) + k_i e_t \Delta t + u_{t-1},
\]
where the constants $k_p, k_i$ are parameters that we will use \ac{RL} to determine.
We adopt \iac{PI} controller structure for two reasons: 
such configurations are widely used in practice,
and even this simple structure can illustrate the challenges associated with stability while achieving excellent performance in \ac{RL} tasks.
\par

We run two \ac{RL}-based experiments: 
one with no stability constraint and
one where stability is enforced by a projection-based update scheme.
For the second, we project the parameter vector $\hat{\theta} = [k_p, k_i]$ proposed by the RL algorithm by solving the optimization problem below:
\begin{equation}
\begin{aligned}
    &\underset{\theta}{\text{minimize}} && \norm{\theta - \hat{\theta}}\\
    &\text{subject to} && \pi_{\text{PI},\theta} \in \mathcal{K}_{\text{stable}}.
\end{aligned}
\label{eq:PIDprojection}
\end{equation}

\begin{figure}[tbp]
\begin{center}
\includegraphics[width=8.4cm]{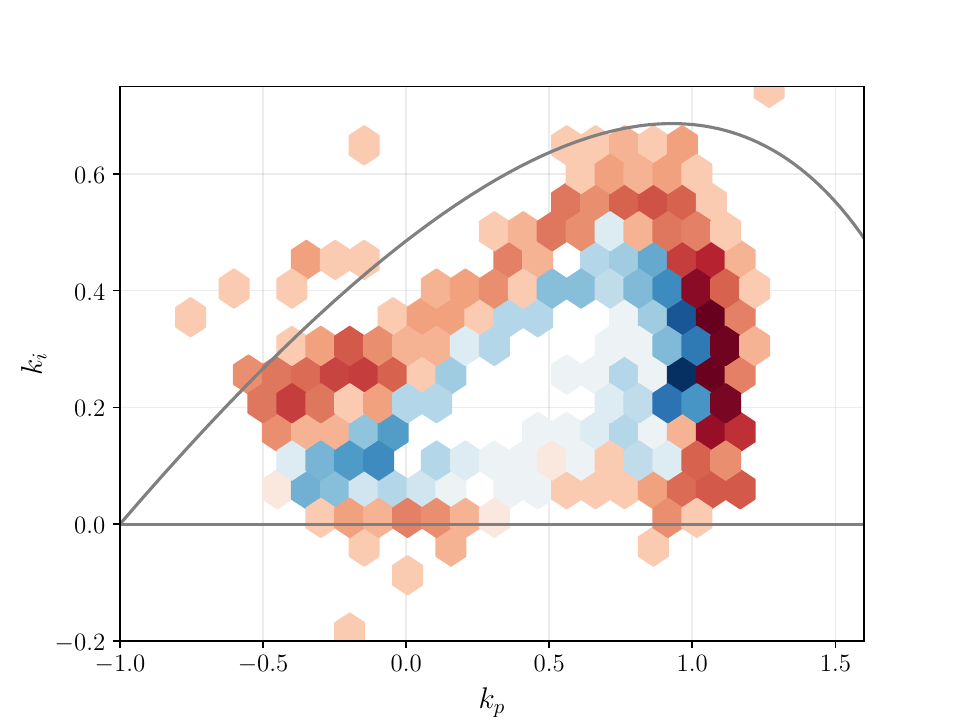}    
\caption{Heatmap of projected \acs{PI} parameters strictly inside the stability boundary.}
\label{fig:PIDparams}
\end{center}
\end{figure}

\Cref{fig:rewards_PID} illustrates the training performance of the two experiments. We implemented a sparse reward function by defining $r(s_t) = 1$ if $\abs{e_t} < \delta$ and $0$ otherwise, where $\delta$ is a small constant. Moreover, we ran the \ac{RL} algorithm $10$ times for each experiment using the default hyperparameters. We actually tweaked the actor learning rate for the unconstrained experiment to make the results more competitive.
Although it is possible to improve the unconstrained results through trial and error, this underscores the importance of stability-based methods.
Imposing the minimal intervention in \cref{eq:PIDprojection} avoids the dangerous, low-reward regions altogether.
\par

\Cref{fig:PIDparams} accompanies the constrained experiment that produced \cref{fig:rewards_PID}. It shows the distribution of \ac{PI} parameters over the $10$ training sessions. Specifically, we depict pre-projection (red) and post-projection (blue) values. (We removed parameter values that did not move substantially to avoid mixing the red and blue regions.)
The gray curve shows the stability boundary for the underlying system. We avoided overlapping the blue region with the boundary by constraining the maximum eigenvalue of the optimization variables $X, Y, W$ in \cref{eq:affineYK}.
In contrast, the parameters corresponding to the unconstrained experiments in \cref{fig:rewards_PID} (\emph{not} the red values in \cref{fig:PIDparams}) can leave the interior, then either recover automatically or not at all.

\section{Discussion and conclusions}
\label{sec:conclusion}

\subsection{Extension to MIMO and unstable systems}
\label{subsec:extension}

The ingredients put forth here can, in principle, handle \ac{MIMO} and unstable systems. Willems' fundamental lemma applies for \ac{LTI} systems, regardless of dimension, and makes no claims about stability.
The \ac{YK} parameterization, as presented in \cref{subsec:dataYK}, also applies to \ac{MIMO} systems. However, the intuitive derivation given there does not apply to unstable systems.
Nonetheless, the most straightforward approach is to apply the constraint-based characterization of stabilizing controllers due to \citet{furieri2019InputOutputParametrization} utilized in \cref{subsec:exPIDtune}. (This is in contrast to the ``classical'' \acs{YK} approach of factorizing the plant $P$.)
Such an approach does not require the plant to be stable, but the controller is no longer characterized in closed form using a single free parameter $Q$.

The constraint-based approach can be used to obtain an initial stabilizing controller.
This controller can then be refined using \ac{RL}, then follow the projection strategy detailed in \cref{subsec:exPIDtune}. 
Alternatively, one may opt for a data-driven control strategy. To apply the \ac{YK} parameterization in an analogous way to that in \cref{subsec:dataYK}, one may then augment the output of the initial stabilizing controller by adding the $Q$ parameter, such as in \cref{subsec:extank}.
One confounding factor in both of these approaches is collecting appropriate data from an unstable system and reliably generating rollouts from the Hankel-based model.
This requires extending the results in \cref{sec:stableHankel} and is a promising avenue for future work.

\subsection{Conclusion}
The \ac{YK} parameterization is well-known in control theory but seemingly under-utilized in \ac{RL}. Taking it as a starting point, we have adapted advances in deep learning and behavioral systems to develop an end-to-end framework for learning stabilizing policies with general \ac{RL} algorithms. 
These core ingredients invite a modular approach to learning stabilizing controllers in which past, present, and future components are cross-compatible. For example, the nonlinear $Q$ parameterization in \cref{sec:Qparam} is \emph{functional}, rather than \emph{structural}: as long as smoothness and the Lyapunov hypotheses are satisfied, one has freedom in terms of activations, layers, or architecture altogether.
Alternatively, one may also elect to use ``classical'' approaches---simpler learning algorithms, restricted sets of linear operators, or observer-based control instead of employing Willems' lemma as an internal model---in combination with newer ones.

There are many further avenues to explore. These include the use of stochastic policies, extensions to unstable systems, and balancing the persistence of excitation assumption during training and steady-state operations. We believe this is a fruitful area to investigate further as deep \ac{RL} gains traction in process systems engineering.

\section*{Acknowledgement}
\label{sec:acknowledgements}
We gratefully acknowledge the financial support of the Natural Sciences and Engineering Research Council of Canada (NSERC) and Honeywell Connected Plant. We would also like to thank Professor Yaniv Plan for helpful discussions.

\renewcommand*{\bibfont}{\footnotesize}
\bibliographystyle{elsarticle-num-names}
\bibliography{autosam.bib}           
\appendix
\section{Implementation details}
\label{app:implementation}

Numerical experiments were carried out in the Julia programming language. 
We utilized \texttt{ReinforcementLearning.jl} \citep{Tian2020Reinforcement}, \texttt{ControlSystems.jl} \citep{carlson2021controlsystems}, and \texttt{NLOpt.jl} \citep{legat2022mathoptinterface}.

As discussed in \cref{sec:stableRL}, any \ac{RL} algorithm may be employed as long as the user provides an appropriate $Q$ parameterization to represent the policy. 
For approaches based on random search or direct methods, one may simply generate rollouts via \cref{alg:YK} inside an optimization program.
However, this strong dependence between rollouts and policy parameters can break when using policy gradient-based methods. We take the \ac{LTI} case in \cref{eq:QLTI} as an example. If one stores $z_t$ and  $e_t + \bar{y}_L$ as the \ac{RL} state, then training the policy---that is, the $Q$ parameter---as $\pi(s_t) = C_q z_t + D_q(e_t + \bar{y}_L)$ will not result in updates to the $A_q$ and $B_q$ matrices. Therefore, even though the environment can be rolled out with \cref{eq:QLTI}, the policy requires $z_t$ to \emph{explicitly} be a function of $A_q$ and $B_q$, namely, by unrolling \cref{eq:QLTI} for one time step. Once the policy is written in an appropriate fashion, policy gradient-based \ac{RL} implementations can automatically compute each gradient component $\frac{\partial \pi}{\partial \theta_i}$ with $\theta$ being a vector of all components in $A_q, B_q, C_q, D_q$.

\section{Further details on random Hankel matrices}
\label{app:randomHankel}

\minsigma*

\begin{proof}
First consider some realization of $\mathcal{H}$ and specific values of $L,N$,
with $N\ge L$.
Simplify notation by writing $H=H_{L,N}$ for the specific $L\times N$ 
Hankel matrix of interest,
and let its rows define the $N$-component vectors
\[
\bar\omega_i = \bbmat \omega_i & \omega_{i+1} & \cdots & \omega_{i+N-1} \ebmat,
\qquad i=0,2,\ldots,L-1.
\label{eq:defobar}\]
Then $\sigma_{\rm min}(H)^2$ is the smallest eigenvalue of the $L\times L$ matrix $H\HT$.

To estimate this minimum eigenvalue, we split the matrix of interest as $H\HT=D+R$, 
where $D$ is the diagonal and $R=H\HT-D$ is the remainder. (It is helpful to write down these matrices in terms of \cref{eq:defobar} for reference.)
$H H\transpose $ is symmetric, so the variational characterization of eigenvalues gives
\[
\lambda_{\min}(H\HT) 
= \lambda_{\min}(D+R) 
\ge
\lambda_{\min}(D) + \lambda_{\min}(R).
\]
We expect that $\lambda_{\min}(D)$ is ``large'', and $\abs{\lambda_{\min}(R)}$ is ``small''.
Let us quantify these intuitions under two preliminary conditions.
Assume first that some fixed real parameter $\theta$ dominates
the magnitude of every entry in $R$, that is,
\[
\abs{\inner{\bar\omega_i}{\bar\omega_j}} \le \theta N,
\qquad \forall i,j\in[0,L-1]\ \text{with}\ i\ne j.
\label{eq:smallproducts}\]
Next, assume that some $\alpha\in(0,1)$ obeys
\[
\omega_{L-1}^2 + \ldots + \omega_{N-1}^2
\ge \alpha(N-L+1).
\label{eq:bigsquares}\]

For the matrix $R$, 
Gershgorin's Circle Theorem implies
\[
\lambda_{\rm min}(R) 
\ge \min_i \left(- \sum_{j\ne i}\abs{\inner{\bar\omega_i}{\bar\omega_j}} \right)
\ge -N\theta(L-1).
\]

For the matrix $D$, each diagonal entry is a sum of $N$ squares.
Every such sum includes the $N-L+1$ terms on the left side of~\cref{eq:bigsquares}.
Thus \cref{eq:bigsquares} provides a lower bound for every diagonal entry in $D$,
and of course one of those diagonal entries is the smallest.
We deduce that
\[
\lambda_{\min}(D)
\ge \omega_{L-1}^2 + \ldots + \omega_{N-1}^2
\ge \alpha(N-L+1).
\]
We conclude that
\[
\lambda_{\rm min}(H\HT)
\ge \alpha(N-L+1) - N\theta(L-1).
\label{eq:oneLoneN}\]
With the specific choices
\[
\theta = \frac{1}{L+1},\qquad
\alpha = \frac{L}{L+1},
\]
we have both $\alpha,\theta\in(0,1)$ and $\alpha=L\theta$, leading to
\[
\lambda_{\rm min}(H\HT)
\ge N\theta - \theta L(L-1)
= \frac{N}{L+1} - \frac{L(L-1)}{L+1}.
\label{eq:lambdamin}\]
Define $r_N>0$ by matching $r_N^2$ with the right side here.
Then $\sigma_{\rm min}(H)\ge r_N$.

Continuing with fixed $N$ and $L$,
let us now estimate the probabilities 
of the prerequisite inequalities above.
In condition~\cref{eq:smallproducts}, the inner product fits
the pattern in~\cref{cor:HW}(b), and we have
\[
\Pr\set{\abs{\inner{\bar\omega_i}{\bar\omega_j}}\le N\theta}
\ge 1 - 2\exp\left(-c_1\theta^2 N\right).
\]
For condition~\cref{eq:bigsquares}, \cref{cor:HW}(a) gives
\[
\begin{multlined}
\Pr\set{ \omega_{L-1}^2 + \ldots + \omega_{N-1}^2
\ge \alpha(N-L+1) } \\
\ge 1 - 2 \exp\left(-c_0 \alpha^2 (N - L + 1)\right).
\end{multlined}
\]
As $N\to\infty$, the $L(L-1)/2$ events in~\cref{eq:smallproducts}
and the further condition in~\cref{eq:bigsquares} have probabilities
that converge to $1$ exponentially quickly.
The same must be true of their intersection, 
and we have shown that this covers the situation where $\sigma_{\rm min}(H)\ge r_N$.
This completes the proof.
\end{proof}

\HplusHprime*

\begin{proof}
Let us write $\omega'=(\omega_{N-L},\ldots,\omega_{N-1})$ for
the last column in $\bar H$, to create the block-structured
expression $\bar{H} = [H\ \omega']$.
Then, using cyclic permutation and the upper bound in \cref{eq:spectralupperbound},
\begin{align}
\rho\left(H^+ H'\right)^2 
&\le \rho\left(\bar{H}\transpose  \left( H H\transpose  \right)^{-1} \bar{H}\right) \\
&= \rho\left( \left( H H\transpose  \right)^{-1} \bar H\, \bar{H}\transpose  \right) 
\\
&= 1 + \rho\left( \left( H H\transpose  \right)^{-1} \omega' {\omega'}\transpose  \right).
\end{align}
Here the final equation holds because the matrix added to $I$ in the line above is positive semi-definite, 
being the product of two factors that are each positive semi-definite and symmetric.
Indeed, these same two properties support the following estimate:

\begin{align}
\rho\left( \left( H H\transpose  \right)^{-1} \omega' {\omega'}\transpose  \right) 
&\leq
 \norm{\left( H H\transpose  \right)^{-1}} 
 \norm{\omega' {\omega'}\transpose }\\
&= \frac{\rho( \omega' {\omega'}\transpose  )}{\min_{x \neq 0} \frac{\norm{H H\transpose  x}}{\norm{x}}} 
\\
&= \frac{ {\omega'}\transpose  \omega' }{\lambda_{\min}\left( H H\transpose  \right)}.
\end{align}

We arrive at the intermediate result
\[
\rho(H^+H')
\le
\sigma_{\max}\left( H^+ H' \right) 
\leq \sqrt{1 + \frac{\norm{ \omega' }^2}{\lambda_{\min}\left( H\HT \right)} }.
\label{eq:stablewall}
\]
Here \cref{prop:minsigma} is relevant.
Let $\set{r_N}$ be a sequence with $r_N\to\infty$ for which
\[
\Pr\set{\lambda_{\rm min}(H\HT)\ge r_N} \to 1.
\label{eq:usetheprop}
\]
Invent any sequence $\delta_N$ with $\delta_N\to 0$ such that $r_N\delta_N \to \infty$.
Manipulate random events as follows:
\begin{align}
\set{ \frac{\norm{ \omega' }^2}{\lambda_{\rm min}(H\HT)} > \delta_N }
&= \begin{multlined}\set{ \frac{\norm{ \omega' }^2}{\lambda_{\rm min}(H\HT)} > \delta_N }\\
   \cap\left( \set{\lambda_{\rm min}(H\HT)\le r_N} \cup \set{\lambda_{\rm min}(H\HT)> r_N}\right)\end{multlined}
\\
&\subseteq \set{\lambda_{\rm min}(H\HT)\le r_N} \cup \set{\frac{\norm{ \omega' }^2}{r_N} > \delta_N}.
\end{align}

The first event on the right is controlled by \cref{eq:usetheprop},
while Markov's inequality gives
\[
\Pr\set{\frac{\norm{ \omega' }^2}{r_N} > \delta_N}
\le \frac{\EE\norm{ \omega' }^2}{\delta_N r_N}
=  \frac{L^2}{\delta_N r_N}
\to 0.
\]
In view of \cref{eq:stablewall}, we have
\[
\Pr\set{ \sigma_{\max}\left( H^+ H' \right) > \sqrt{1 + \delta_N} }
\to 0\qquad\text{as}\ N\to\infty.
\]
The stated result is an elementary reformulation of this.
\end{proof}

\end{document}